\documentclass{article}

\usepackage[nonatbib, final]{neurips_2019}

\usepackage[utf8]{inputenc} 
\usepackage[T1]{fontenc}    
\usepackage{hyperref}       
\usepackage{url}            
\usepackage{booktabs}       
\usepackage{amsfonts}       
\usepackage{nicefrac}       
\usepackage{microtype}      
\usepackage{amsmath}
\usepackage{amssymb}
\usepackage{amsfonts}
\usepackage{algorithm}
\usepackage{algorithmic}
\usepackage{caption}
\usepackage{graphicx}
\usepackage{color}
\usepackage{hyperref}
\usepackage{parskip}
\usepackage{bm}
\usepackage{float}
\usepackage[table,x11names]{xcolor}
\usepackage{multirow}

\newcommand\ml[1]{\multicolumn{1}{|l|}{#1}}
\newcommand\mcc[1]{\multicolumn{2}{c}{#1}}
\DeclareMathOperator*{\argmax}{argmax} 
\DeclareMathOperator*{\argmin}{argmin} 

\title{Adversarial Examples for Cost-Sensitive Classifiers}

\author{%
  Gavin S. Hartnett
  \\
  The RAND Corporation\\
  \texttt{hartnett@rand.org} \\
  \And
  Andrew J. Lohn \\
  The RAND Corporation\\
  \texttt{alohn@rand.org} \\
  \And
  Alexander P. Sedlack \\
  The RAND Corporation\\
  \texttt{asedlack@rand.org} \\
}

\begin{document}

\maketitle

\begin{abstract}
Motivated by safety-critical classification problems, we investigate adversarial attacks against cost-sensitive classifiers. We use current state-of-the-art adversarially-resistant neural network classifiers \cite{xie2018feature} as the underlying models. Cost-sensitive predictions are then achieved via a final processing step in the feed-forward evaluation of the network. We evaluate the effectiveness of cost-sensitive classifiers against a variety of attacks and we introduce a new cost-sensitive attack which performs better than targeted attacks in some cases. We also explored the measures a defender can take in order to limit their vulnerability to these attacks. This attacker/defender scenario is naturally framed as a two-player zero-sum finite game which we analyze using game theory.
\end{abstract}

\section{Introduction}
Many safety-critical classification problems are not indifferent to the different types of possible errors. For example, when classifying tumors in medical images one may be relatively indifferent between misclassifications within the super-categories of malignant or benign tumors, but may be particularly interested in avoiding misclassifications across those categories, for example misidentifying a malignant Lobular Carcinoma tumor as being instead a benign Fibroadenoma tumor \cite{cancerClassifier}. 

It is a relatively simple matter to adjust the predictions of a trained classifier to reflect the different costs associated with the various types of classification errors using a formalism known as cost-sensitivity \cite{elkan2001foundations, domingos1999metacost}. Without cost-sensitivity, the most likely class is taken to be the prediction made by a classification model. In contrast, cost-sensitive classifiers make predictions by first computing the expected cost associated with each prediction, and then taking the class with the smallest expected cost to be the model prediction.

Motivated by safety-critical classification problems, we investigate adversarial attacks on cost-sensitive classifiers. We use current state-of-the-art adversarially-resistant neural network classifiers \cite{xie2018feature} as the underlying models, and we considered multiple types of attacks, as well as various defensive actions that may be taken to mitigate the effect of the attacks. Our key findings are:
\begin{itemize}
    \item \textbf{Classifiers face a trade-off between maximizing accuracy and minimizing cost:} \\
    Predictions can be made with the goal of either maximizing the accuracy or minimizing the expected cost. While these are not diametrically opposed goals (for example a perfect classifier will incur zero cost), in practice there will be a trade-off where the classifier can make conservative predictions which lower both the cost and the overall accuracy.
    \item \textbf{The attacker faces a trade-off between minimizing accuracy and maximizing cost:} \\
    Similarly, the attacker can craft adversarial examples designed to either minimize the defender's accuracy or increase their average cost. As before, these goals aren't necessarily in conflict with one another - for example if the attacks succeed 100\% of the time, then both goals may be simultaneously accomplished, but in practice attacks will only succeed some fraction of the time and the attacker will be faced with a trade-off.
    \item \textbf{Calibration leads to both better defenses and more effective attacks:} \\
    The expected cost depends on the predicted probabilities of the neural network and not just on the overall class prediction. Therefore, it is important that the classifier produce accurate probability estimates. We find that both cost-sensitive defenses and attacks may be improved by calibrating these estimates.
    \item \textbf{The attacker/defender scenario is naturally analyzed in terms of game theory:} \\
    We explored many different pairings of attacks and defensive measures. The identification of good strategies becomes more difficult as the number of possible scenarios increases. We observe that this problem is naturally framed as a two-player zero-sum finite game, and therefore the game theoretic concepts of Nash equilibria and dominant strategies may be used to analyze the attacker/defender competition.
\end{itemize}


\section{Cost-Sensitivity for classification problems \label{sec:cost-sensitive}}
In this section we provide a brief review of cost-sensitivity \cite{elkan2001foundations, domingos1999metacost}. Throughout this work we shall consider classification problems and denote the inputs as $\bm{x}$, and we will use the indices $A,B$ to run over all possible $K$ classes, i.e. $A \in \{ 1, 2, ..., K\}$.

Of central importance in cost-sensitive classification problems is the cost-matrix $C_{AB}$, which is defined to be the cost of predicting class $A$ when the correct class is $B$. The cost may be measured in any units since the cost-sensitive predictions are unaffected by scaling the cost matrix by an overall constant. We shall require that the costs are non-negative, $C_{AB} \ge 0$, with equality if and only if $A=B$, which reflects the fact that a correct classification should incur no cost. Given the cost matrix and $p(A|\bm{x})$, the model estimate for the probability that an input $\bm{x}$ belongs to class $A$, the expected cost of predicting class $A$ will be denoted as $\mathcal{C}_A(\bm{x})$, and is simply \cite{elkan2001foundations}
\begin{equation} 
\mathcal{C}_A(\bm{x}) = \sum_B C_{AB} \, p(B|\bm{x}) \,.
\end{equation} 
The cost-sensitive (CS) prediction is then the class for which the expected cost is smallest, i.e. 
\begin{equation}
\label{eq:MCprediction}
A_{\text{CS}}(\bm{x}) := \argmin_A \mathcal{C}_A(\bm{x}) \,.
\end{equation}
In contrast, in most classification settings the prediction is taken to be the most likely class:
\begin{equation}
\label{eq:MPprediction}
A_{\text{MP}}(\bm{x}) := \argmax_A p(A|\bm{x}) \,,
\end{equation}
which we shall refer to as the maximum probability (MP) prediction.

\subsection{Geometry of cost-sensitive predictions}
To gain an intuition for how cost-sensitive predictions compare to the more standard maximum probability predictions, it is useful to consider the problem from a geometrical perspective. Binary classification $(K=2)$ is especially simple: if the probability of class 1 is denoted $p$, then the probability of class 2 is $(1-p)$. The maximum probability prediction is determined by whether $p < 0.5$ (prediction is class 2) or $p > 0.5$ (prediction is class 1). The effect of cost-sensitivity then is to shift the decision threshold from $p_* = 0.5$ to a new value determined by the relative cost of the two types of errors. That is, class 1 is predicted if $p > p_*$, where now
\begin{equation}
    p_* = \frac{C_{12}}{C_{12} + C_{21}} \,.
\end{equation}

\begin{figure}
\centering
\begin{minipage}[t]{0.46\textwidth}
  \centering
  \includegraphics[width=0.8\linewidth]{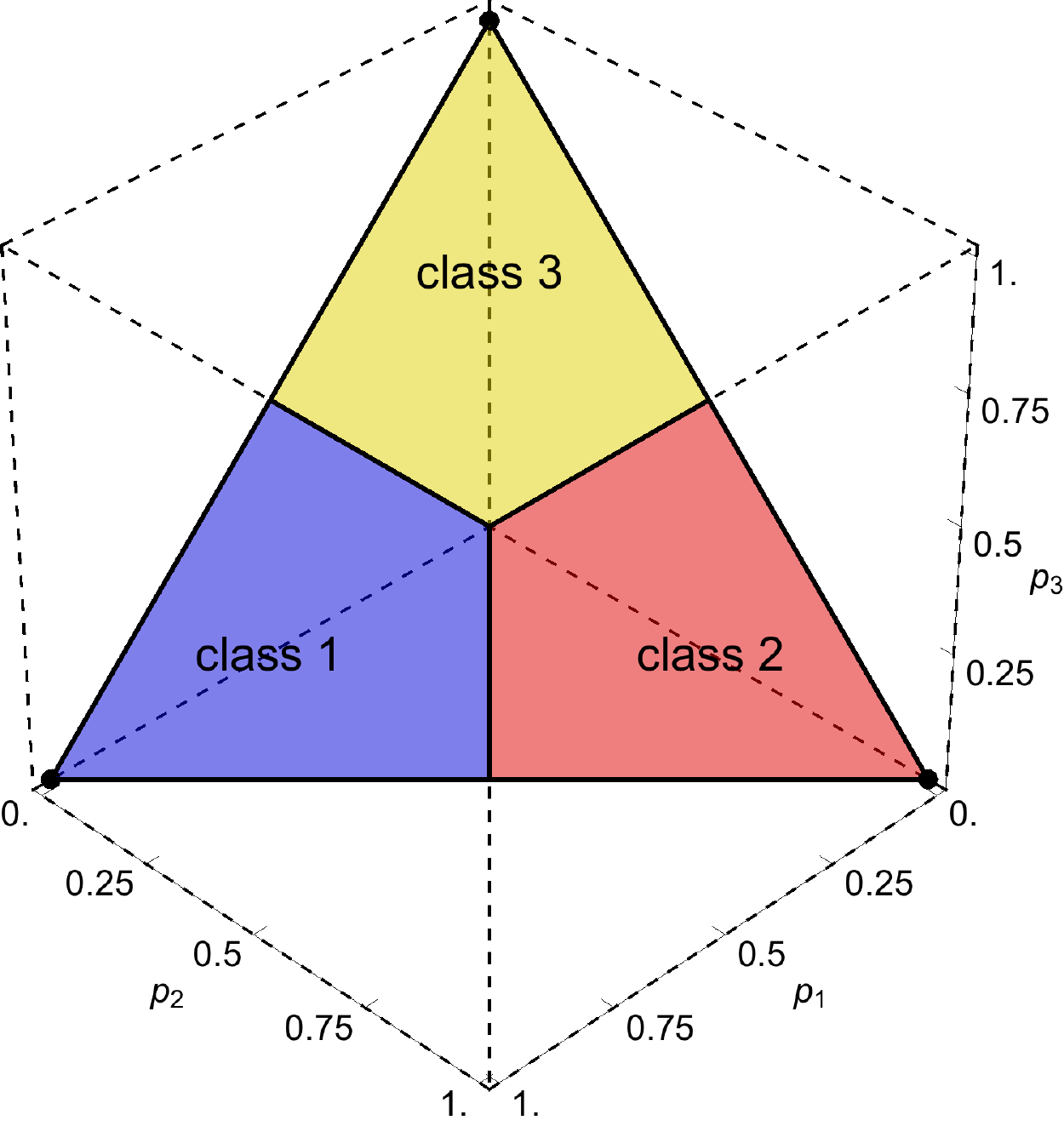}
  \captionof{figure}{The unit 2-simplex as a surface embedded in the 3-dimensional space spanned by the probability coordinates $(p_1, p_2, p_3)$. The simplex has been divided into cells corresponding to the maximum-probability prediction. The figure has been oriented with the origin behind the simplex.}
  \label{fig:maxprob}
\end{minipage}%
  \hfill
\begin{minipage}[t]{0.46\textwidth}
  \centering
  \includegraphics[width=0.8\linewidth]{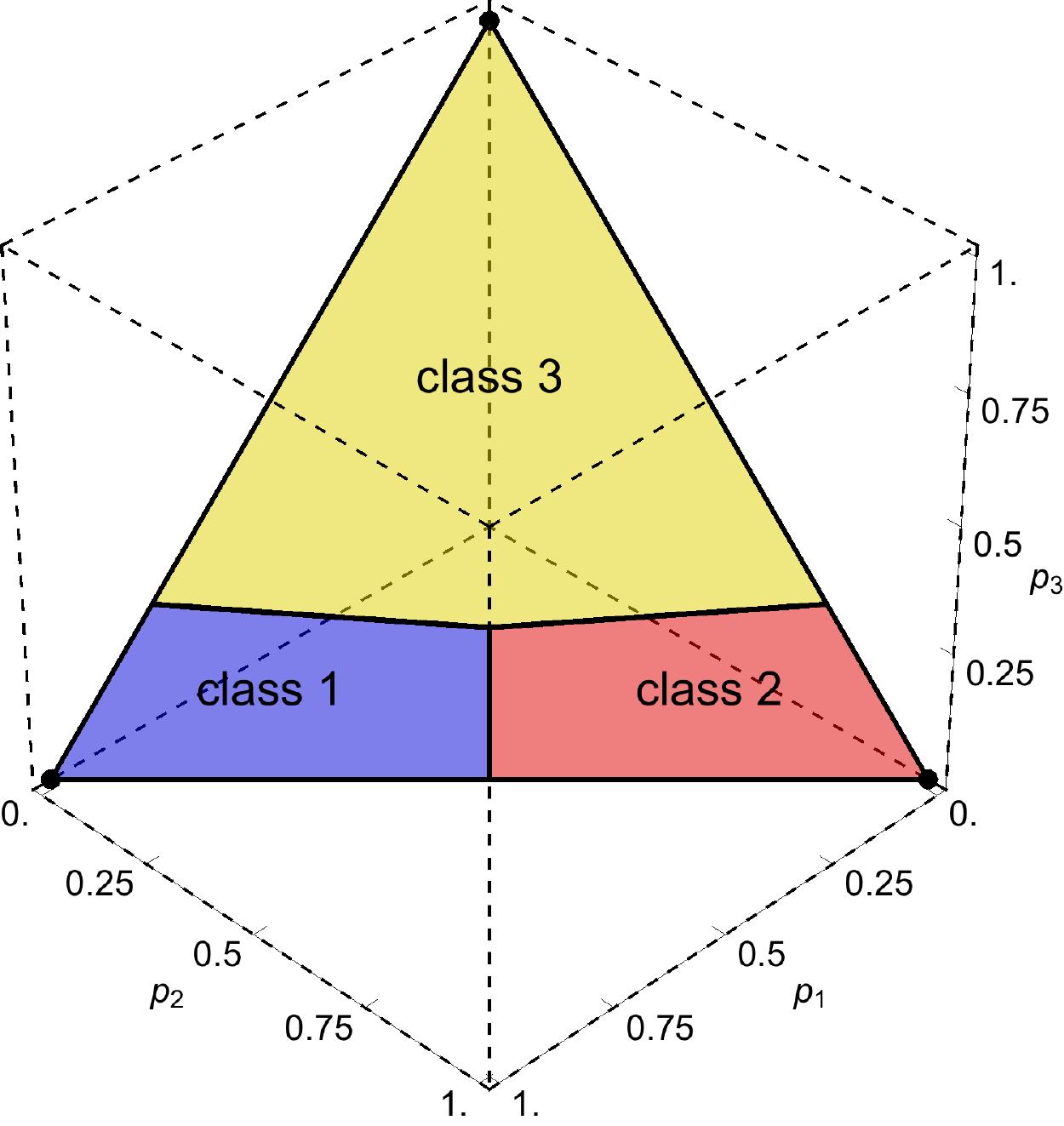}
  \captionof{figure}{The same simplex plotted in Fig.~\ref{fig:maxprob}, now with the cells determined according to the minimum cost prediction. The cost matrix used here is given by $C_{12} = C_{21} = 1$, $C_{31} = C_{32} = 3$, and $C_{13} = C_{23} = 10$  (and zero diagonal entries). Misidentifying class 3 incurs a large cost, and hence the class 3 cell has expanded accordingly.}
  \label{fig:mincost}
\end{minipage}
\end{figure}

The higher dimensional case of $K \ge 3$ is more interesting. In this case it is more useful to work in terms of the probability simplex, which is a $(K-1)$-dimensional hyper-surface embedded in $K$-dimensions. The embedding coordinates are the class probabilities, i.e. $p(A|\bm{x})$ for an input $\bm{x}$, and the simplex is the surface satisfying the constraints $0 \le p(A|\bm{x}) \le 1$ $\forall A$ and $\sum_A p(A|\bm{x}) = 1$. The vertices of the simplex are the points for which all the probability mass is placed on a single class. The simplex may be divided into cells such that all points within a single cell will lead to the same maximum probability prediction. This is depicted for $K=3$ in Fig.~\ref{fig:maxprob}. The effect of cost-sensitivity is to shift the cell boundaries, for example as in Fig.~\ref{fig:mincost}. In general, cells representing classes which are costly to misidentify, for example the malignant Lobular Carcinoma tumor discussed above, will expand, corresponding to an increased risk aversion.

\subsection{Multi-class classification problems with two super-categories} 
A general cost-matrix for $K$-class classification is determined by $K(K-1)$ parameters (assuming that the diagonals are zero, representing zero cost for correct predictions). Both for simplicity and because we are motivated by scenarios such as the benign/malignant tumor classification discussed above, we consider a much smaller family of cost-matrices. We split the $K$ classes into 2 super-categories, which we call the ``sensitive" and ``insensitive" categories. Let there be $m$ members of the insensitive group, and $K-m$ members of the sensitive group, and split the label index $A$ as $a = 1,...,m$ for the insensitive group members, and $\alpha = m+1,...,K$ for the sensitive group. We shall consider scenarios where the main concern is inter-category misclassifications, especially misclassifying a sensitive class as an insensitive class (i.e. misidentifying a malignant tumor as a benign tumor). Intra-category misclassifications will also have associated costs, albeit they will be less significant than inter-category costs.

In this scenario, we can break the cost-matrix into 4 blocks,
\begin{equation}
C_{AB} = 
\begin{pmatrix}
    C_{a b} & C_{a \beta} \\
    C_{\alpha b} & C_{\alpha \beta}
  \end{pmatrix} \,,
\end{equation}
and take each constituent block matrix to be
\begin{equation}
C_{ab} = c^{(ab)} \left(1-\delta_{ab}\right) \,, \quad C_{a\beta} = c^{(a\beta)} \,, \quad C_{\alpha b} = c^{(\alpha b)} \,, \quad C_{\alpha \beta} = c^{(\alpha\beta)}\left(1 - \delta_{\alpha \beta} \right) \,.
\end{equation}
Here the lower-case $c$'s are constants, and $\delta_{ab}$ and $\delta_{\alpha \beta}$ are Kronecker deltas. The constant $c^{(ab)}$ is the cost of misclassifications within the insensitive super-category, and $c^{(\alpha \beta)}$ is similarly the cost of misclassifications within the sensitive super-category. The off-diagonal term $c^{(a\beta)}$ represents the cost of mis-labeling a sensitive class as insensitive, and vice versa for $c^{(\alpha b)}$. Motivated by safety-critical scenarios where the most costly type of mistake is mis-identifying a sensitive class as insensitive, we will assume that the different costs obey the following inequalities:
\begin{equation} 
0 < c^{(ab)} \le c^{(\alpha \beta)} \le c^{(\alpha b)} \ll c^{(a \beta)} \,,
\end{equation}
so that the cost-matrix is determined by just 4 independent numbers.

\section{Adversarial examples for cost-sensitive classifiers \label{sec:advexamples}}
Cost-sensitivity is particularly relevant for safety-critical scenarios because it enables classifiers to take into account the fact that some mistakes are more deleterious than others. This general framework naturally complements the context of adversarial examples, which are artificially generated inputs of a classifier designed to cause mistakes \cite{szegedy2013intriguing}, and which are an important threat for safety-critical applications of classifiers.\footnote{See \cite{gilmer2018motivating} for an analysis of the concrete ways in which adversarial examples are relevant for AI Safety.} The general idea that different misclassifications are associated with different costs should be reflected both in how the classifier makes predictions and in the types of adversarial attacks a malicious actor would choose to employ in order to cause maximum damage.

Concretely, we consider an attacker/defender scenario in which the defender is a neural network classifier, and the attacker is an agent attempting to fool the defender by presenting it adversarial examples. We will investigate multiple types of attacks against classifiers making predictions according to both the maximum probability criterion and the minimum cost criterion. We also consider both white-box and black-box scenarios, in which the attacker has or does not have access to the defender network, respectively. 

As usual, we take the attack to be defined by a constrained  optimization problem. To set notation, let $\ell(\bm{x})$ be the objective function ($\ell$ may also depend on other quantities such as the target label), and the optimization problem is then
\begin{equation}
\label{eq:targetedattack}
\bm{\delta} = \argmax_{\bm{\delta} \in \Delta} \ell\left(\bm{x} + \bm{\delta} \right) \,.
\end{equation}
Here $\Delta$ is the attack set, the set of allowable perturbations around a given clean input $\bm{x}$. Throughout this work, we will take $\Delta$ to be an $\epsilon$-ball in the $\ell_{\infty}$ norm, i.e. $\Delta = \{ \bm{\delta} : ||\bm{\delta}||_{\infty} \le \epsilon \}$.\footnote{Other authors, most notably \cite{gilmer2018motivating}, have noted a number of shortcomings in using this attack set for research into safety-critical implications of adversarial examples. We do not disagree with these observations, but will work with the $\ell_{\infty}$ ball nonetheless both for mathematical convenience and because we regard this issue as orthogonal to the main idea of the current work, which is the relevance of cost-sensitivity to adversarial example research.} Independent of the objective function, in all cases we shall use the same projected gradient descent (PGD) method of \cite{madry2017towards} to solve the optimization problem and to generate examples. The attack PGD update rule is
\begin{equation}
\bm{x}^{(t+1)} = \Pi_{\bm{\Delta}} \left[ \bm{x}^{(t)} + \alpha \, \text{sign} \left(\nabla_x \, \ell(\bm{x}^{(t)}\right) \right] \,,
\end{equation}
where $\bm{x}^{(t)} := \bm{x} + \bm{\delta}^{(t)}$ represents a sequence of perturbed inputs, $\alpha$ is the step-size parameter, and $\Pi_{\bm{\Delta}}$ is a projection operator that projects the perturbation down to the attack set $\Delta$. The initial perturbation, $\bm{\delta}^{(0)}$, will be randomly initialized within the attack set $\Delta$. 

\subsection{Targeted attacks}
We will consider two types of adversarial attacks. The first is a targeted attack, where the objective function is given by the negative cross-entropy of the target label. That is, if $A_{\text{target}}$ is the target label, then 
\begin{equation}
    \ell_{\text{CE}}(\bm{x}, A_{\text{target}}) =  \log p(A_{\text{target}} | \bm{x} + \bm{\delta}) \,.
\end{equation}
In terms of the probability simplex coordinates, the optimal solution is when all the probability mass has been placed on the target class, i.e. $p(A_{\text{target}}|\bm{x}) = 1$. The target class could be chosen randomly, or it could be chosen to induce a particularly costly error. As an example in the cost-sensitive setting, an effective attack would be one which tricked the classifier into thinking that $\bm{x}$ belonged to the insensitive class when in fact it belonged to a sensitive one.

\subsection{Maximum minimum expected cost attacks}
If the goal of the attacker is to increase the costs of the defender's mistakes, it is natural to consider an attack which is designed to explicitly increase the expected cost. Therefore, we introduce the Maximum Minimum Expected Cost Attack (or maxi-min attack for short): 
\begin{equation}
\label{eq:maximin}
\ell_{\text{maxi-min}}(\bm{x}) = \min_A \mathcal{C}_A(\bm{x} + \bm{\delta}) \,.
\end{equation}

Unlike the targeted attack, the maxi-min attack does not depend on the true class. Thus, the maxi-min attack always aims to modify the input $\bm{x}$ so that the point in the probability simplex $p(A|\bm{x}+\bm{\delta})$ moves to the point of maximal $\min_A \mathcal{C}_A(\bm{x} + \bm{\delta})$, which by symmetry can be seen to be the intersection point where the costs are identical for all class predictions $A$. In particular, for the $K=3$ example of Fig.~\ref{fig:mincost}, this is the point where all 3 cell boundaries intersect. Because this attack aims to bring $p(A|\bm{x}+\bm{\delta})$ to an interior point in the simplex, as opposed to a vertex, it will not be as effective as a targeted attack with cost-sensitive targets - assuming that the optimization problem associated with both attacks can be fully solved. For example, for the cost matrix considered in Fig.~\ref{fig:mincost}, the expected cost at the intersection of all three cell boundaries is $\mathcal{C}_A = 12/5$, whereas a cost of $C_{13} = 10$ could be achieved if the prediction was 1 and the true class was 3. However, the optimization problem defining adversarial attacks is rarely able to be solved exactly, and thus there could well be instances where the maxi-min attack is more effective - indeed, we shall find this to be the case in what follows.

As far as we are aware, we are the first to consider adversarial attacks designed to directly maximize the cost. Recently, Zhang and Evans \cite{zhang2018cost} considered a cost-sensitive extension of Wong and Kolter's approach towards developing provably robust classifiers \cite{wong2017provable}. In the Zhang and Evans extension, robustness is defined with respect to cost, as opposed to the overall misclassification error. Our work is complementary to theirs as we consider attacks designed to explicitly increase the cost. 

\section{Attack comparison \label{sec:attackcomparison}}
In this section we detail the numerical experiments used to compare the efficacy of the 3 different types of attacks considered here.

\subsection{Experimental set-up}
We considered the task of image classification on the ImageNet dataset \cite{ILSVRC15}. Our motivating interest is near-term scenarios in which an imperfect but high-performance image classification system is employed in a safety-critical application. Given the amount of attention adversarial examples have received, it seems plausible that many organizations will be cognizant of the threat posed by adversarial examples, and will therefore choose to employ models with some level of resistance. For simple enough problems one can obtain provable guarantees regarding robustness (see for example \cite{wong2017provable, raghunathan2018certified} and references therein), but these methods do not currently scale for modern image classifiers trained on high-resolution images.\footnote{As this work was nearing completion progress on this problem was made in \cite{cohen2019certified}.} Thus, we shall focus on problems for which the vulnerability to adversarial examples can only be mitigated, not fully eliminated or bounded. 

We consider attacking networks which have been adversarially trained \cite{goodfellow2014explaining, kurakin2016adversarial, kannan2018adversarial, madry2017towards}, so that they are somewhat resistant to adversarial attacks. In particular, we used pre-trained models released as part of the recent work \cite{xie2018feature}. Three such pre-trained models were released: ResNeXt-101, ResNet-152 Denoise, and ResNet-152 Baseline. These models obtain between 62-68\% top-1 accuracy on clean images, and 52-57\% accuracy on adversarially perturbed images with random targets (we specify the attack details below). All three models were trained on adversarial examples, and the first two also incorporate a novel form of feature de-noising to enhance their resistance to adversarial examples.

A simple but crucial point is that a cost matrix is required in order to implement cost-sensitive predictions. The cost matrix encapsulates the costs associated with different types of mistakes, but these may be hard to quantify in certain applications. To return to the example of identifying malignant tumors, clearly false positives are less costly mistakes than false negatives, but are they 10x worse, 100x worse, or 1000x worse? These valuations must be made for each application, and could involve a rich set of considerations which we shall not get into here. Instead, we simply consider an arbitrary cost matrix with values chosen according to what seems like plausible values. In particular, we let there be $m=900$ insensitive classes and $K-m=100$ sensitive classes, with the costs taken to be\footnote{We note that we randomly permuted the ImageNet labels in order to avoid grouping together similar classes in the insensitive/sensitive super-categories.}
\begin{equation}
\label{eq:costs}
    c^{(ab)} = 1\,, \qquad 
    c^{(\alpha \beta)} = 2 \,, \qquad
    c^{(\alpha b)} = 5 \,, \qquad 
    c^{(a\beta)} = 200 \,.
\end{equation} 

Although these values were mostly chosen arbitrarily, they were picked so that the effect of being cost-sensitive would be non-trivial. For example, as $c^{(a\beta)} \rightarrow \infty$, with the other values held constant, a cost-sensitive classifier will always err on the side of caution and predict the sensitive class. Similarly, if the differences in cost are very slight, then a cost-sensitive classifier will mostly make predictions according to the most likely class. These values were chosen to avoid either extreme. An additional complication is that an adversary may not know (or may only partially know) the cost matrix used by the defender network. Thus, in cost-sensitive adversarial examples the cost-matrix becomes part of the white-box/black-box characterization of the problem. In this work, we assume that the cost matrix is known to the attacker.


\subsection{Experimental results \label{sec:experimental_results}}
We generated adversarial attacks using the ResNeX1-101 pretrained model of Ref.~\cite{xie2018feature}, and evaluated the attacks against each of the 3 pretrained models. The attack is a white-box attack when the defending network is the same ResNeX1-101 model used to generate the attacks, and it is a black-box attack when the defending network is either of the ResNet-152 models. We considered 3 types of attacks: targeted with random targets, targeted with cost-sensitive targets, and the maxi-min attack introduced in Sec.~\ref{sec:advexamples}. We use the same attack parameters as in \cite{xie2018feature}, and used PGD to generate attacks for $n_{\text{steps}}$ numbers of steps. The attacks are constrained to lie in an $\ell_{\infty}$ ball with $\epsilon = 16$, and the step-size was taken to be $\alpha = 1$ (except for the case $n_{\text{steps}} = 10$, in which case we set $\alpha=1.6$). Furthermore, each attack was randomly initialized in the $\ell_{\infty}$ ball.

In Table~\ref{table:ImageNetresultsWhiteBox} we present the results for white-box attacks generated using the ResNeXt-101 model. The attack details are as follows. The number of PGD iterations was taken to be $n_{\text{steps}} = 10$, and the results in this table were computed by averaging over 50,000 distinct attacks, one for each of the images in the ImageNet validation set. Both the accuracy and average cost are evaluated for the two prediction methods discussed above, maximum probability and minimum cost. The column abbreviations are MP Acc - maximum probability prediction accuracy, MP Cost - maximum probability average cost, MC Acc - minimum cost prediction accuracy, MC Cost - minimum cost prediction average cost. The $\pm$ values indicate the 95\% confidence intervals, which were computed by assuming that the means are normally distributed.

\begin{table}[htbp]
\caption{\label{table:ImageNetresultsWhiteBox}White-box attacks}
\centering
\begin{tabular}{ccccc}
    \toprule
    Attack Type & MP Acc. (\%) & MP Cost & MC Acc (\%) & MC Cost \\ \midrule
    \multicolumn{5}{c}{ResNeXt-101} \\ \midrule
    \multicolumn{1}{l}{clean images} & $68.3 \pm 0.4$ & $6.47 \pm 0.30$ & $61.2 \pm 0.4$ & $2.38 \pm 0.13$   \\
    \multicolumn{1}{l}{random targets} & $57.0 \pm 0.4$ & $8.44 \pm 0.34$ & $41.4  \pm 0.4$ & $2.94 \pm 0.08$   \\
    \multicolumn{1}{l}{max cost targets} & $56.9 \pm 0.4$ & $8.45 \pm 0.34$ & $41.0 \pm 0.4$ & $2.98 \pm 0.08$   \\
    \multicolumn{1}{l}{maxi-min cost} & $60.1 \pm 0.4$ & $13.94 \pm 0.44$ & $49.2 \pm 0.4$ & $3.50 \pm 0.16$  \\
    \midrule
\bottomrule
\end{tabular}
\end{table}

There are a number of interesting observations to make. First, it is unsurprising that the accuracy is similar for both types of targeted attacks when the defending network makes maximum probability predictions, since in this case the cost-sensitive targeted attacks represent a fairly large subset of random targeted attacks. However, it is surprising that the cost-sensitive targeted attacks do such a poor job of increasing the cost for both types of predictions. This illustrates that for adversarially-resistant networks such as those of \cite{xie2018feature}, targeted attacks are a poor way to increase the cost. The maxi-min cost attack outperforms all others when it comes to increasing the cost, although it unsurprisingly leads to fewer overall errors. The increase in cost is quite dramatic for a defending network making maximum probability predictions, and although the effect is less significant for minimum cost predictions, it still far outperforms either targeted attack. 

We present additional results for black-box attacks and variable attack strength $n_{\text{steps}}$ in Appendix \ref{sec:uncalibrated}. The black-box attacks performed similarly to the white-box attacks, although they were (predictably) slightly less effective overall. Increasing $n_{\text{steps}}$ significantly improved the performance of the attacks.

\section{Calibration \label{sec:calibration}}
In many machine learning applications, the only output of a classifier that is used is the class prediction. However, there are many scenarios in which the probability estimates $p(A|\bm{x})$ are also used. Cost-sensitive learning is one such example as the minimum cost prediction, Eq.~\ref{eq:MCprediction}, depends upon $p(A|\bm{x})$. A perfect classifier would place all the probability mass on the correct label, i.e. ${p(A|\bm{x}) = \delta_{A, A_{\text{true}}}}$, and the minimum cost prediction would be $A_{\text{MC}} = \argmin_A C_{AA_{\text{true}}} = A_{\text{true}}$.\footnote{Recall that we are assuming that the cost matrix satisfies $C_{AB} \ge 0$, with equality if and only if $A=B$.} For imperfect classifiers, a desirable property of the probability estimates $p(A|\bm{x})$ is that they be \textit{calibrated} \cite{niculescu2005predicting}. A classifier is said to be calibrated if the prediction accuracy agrees with the probability estimates. For example, whenever a calibrated classifier makes a prediction of class $A$ for an input $\bm{x}$ with $p(A|\bm{x})=0.9$, it will be correct on average 90\% of the time. As a result, the probability estimates of calibrated classifiers may be interpreted as confidences.

Both the minimum cost prediction, Eq.~\ref{eq:MCprediction}, and the maximum minimum expected cost attack, Eq.~\ref{eq:maximin}, depend directly on the probability estimates $p(A|\bm{x})$, and so it is natural to wonder if calibration might significantly affect the results, for example by making the minimum cost predictions more robust, or the maximum minimum expected cost attack more effective. Both the attacker and the defender may separately elect to calibrate leading to a total of four possible scenarios. The scenario where neither party calibrates was treated in the previous section, and in Appendix \ref{sec:calibration_scenarios} we present results for remaining scenarios (defender calibrates, attacker calibrates, and both calibrate). We also provide details on the temperature-scaling calibration method used in Appendix \ref{sec:calibration_method}.

\section{Game theoretic analysis \label{sec:gametheory}}
In the above sections and in the appendices we have considered a total of 6 different attacks (targeted with random targets, targeted with cost-sensitive targets, and the maxi-min attack, each of which can be either generated using a calibrated or an uncalibrated network), as well as 4 types of predictions (maximum probability or minimum cost, each of which may be made using a calibrated or an uncalibrated network). A convenient framework for analyzing the resulting 24 possible scenarios is game theory. 

The attacker/defender set-up considered here may be formulated as a finite zero-sum two-player game. The pay-off of the attacker is the average cost, and the defender's pay-off is the negative average cost. The pay-off matrix for this game may be obtained using the uncalibrated results of Table~\ref{table:ImageNetresultsWhiteBox}, together with the calibrated results presented in Table~\ref{table:ImageNetresultsWhiteBox2}, ~\ref{table:ImageNetresultsWhiteBox3}, ~\ref{table:ImageNetresultsWhiteBox4} in Appendix \ref{sec:calibration_scenarios}. Here, $MP$ stands for ``maximum probability", $MC$ for "minimum cost", $RT$ for ``targeted with random targets", $CST$ for ``targeted with cost-sensitive targets", and $MM$ for ``maxi-min". Notice that the first two rows are identical - the temperature scaling calibration method used does not affect the maximum probability prediction, and therefore it also does not affect the average misclassification costs).

\begin{table}[h!]
    \caption{\label{table:payofflarge}Attacker's Pay-off matrix}
    \begin{center}
    \renewcommand\arraystretch{1.3}
    \begin{tabular}{lc|c|c|c|c|c|c|}
    \mcc{} & \mcc{Attacker} \\ \cline{3-8}
    & & $RT$, $\neg C$ & $RT$, $C$ & $CST$, $\neg C$ & $CST$, $C$ & $MM$, $\neg C$ & $MM$, $C$ \\ \cline{2-8}
    \multirow{2}{*}{Defender}
    & \ml{$MP$, $\neg C$} & 8.44 & 8.43 & 8.45 & 8.45 & 13.94 & 13.97 \\ \cline{2-8}
    & \ml{$MP$, $C$} & 8.44 & 8.43 & 8.45 & 8.45 & 13.94 & 13.97 \\ \cline{2-8}
    & \ml{$MC$, $\neg C$} & 2.94 & 2.95 & 2.98 & 3.00 & 3.50 & 4.16 \\ \cline{2-8}
    & \ml{$MC$, $C$} & 3.21 & 3.22 & 3.25 & 3.25 & 3.38 & \textbf{3.39} \\ \cline{2-8}
    \end{tabular}
    \end{center}
\end{table}

For this simple game, there is a single pure strategy Nash equilibrium (shown in bold in Table~\ref{table:payofflarge}), which is that the defender makes calibrated minimum cost predictions $(MC, C)$, and the attacker makes calibrated maxi-min attacks $(MM, C)$. Note that $(MM, C)$ is a dominant strategy for the attacker, but $(MC, C)$ is not dominant for the defender.

The result of this simple game theory analysis is that, \textit{in terms of the average cost}, both parties should calibrate, minimum cost predictions are better than maximum probability ones, and the best attack is the maxi-min attack. These conclusions may well change with the many factors that went into this analysis - the cost matrix, the underlying classification problem, the strength of the attacks (measured in terms of $n_{\text{steps}}$ and the size of the attack set $\Delta$), etc. However, this overall framework for comparing strategies should be generally applicable. It is possible that in more complicated scenarios the Nash equilibrium will be a mixed strategy, as opposed to the pure strategy found here.

\section{Conclusions and future directions \label{sec:conclusion}}
Safety critical systems are not likely to operate by simply selecting the most likely outcomes; they will need to consider cost of those outcomes and determine the probability thresholds for their predictions accordingly. At the same time, attacks on these cost-sensitive models are particularly important to study because of the critical nature of these systems. We demonstrated several white-box and black-box attacks on cost-sensitive classifiers built from state-of-the-art adversarially-resistant ResNet image classifiers. These classifiers were made resistant by training them on targeted adversarial examples, and we find that they are still vulnerable to attacks designed to increase the expected cost.


While our experimental results were generated for image classification systems, our general framework should apply more broadly to any classification problem. Cost-sensitive classifiers and attacks thereon can easily be envisioned for text analysis (e.g. be sure not to miss terrorist sentiments) or industrial plant operation (e.g. be sure not to miss irregular signals and alerts that lead to accidents). In fact, most applications are not indifferent between different types of misclassifications, making cost-sensitivity broadly applicable. When those applications are safety-critical, an analysis of the efficacy of attacks and defenses should be carried out. 

Lastly, we conclude with some directions for future work. Much of this work implicitly assumes that both parties (the defender and the attacker) know the cost matrix. In practice, it may be hard to convert an implicit value system based on possibly vague and loosely-shared principles into an explicit numerical matrix. Even when such a task is achievable, there are many scenarios where the attacker would not be expected to have access to this information. Thus, one area of future work involves studying the effect of imperfect knowledge of the cost-matrix for the attacker, and whether the attacker can learn to infer the cost-matrix by observing the classifier predictions (and in turn using this information to construct better attacks). It would also be interesting to study the effect of a noisy cost-matrix, perhaps reflecting the challenges faced by the defender in encoding a value system into a cost matrix.

A second line of work would be to go beyond the pre-trained models of \cite{xie2018feature}, and to consider other forms of adversarially-resistant models, especially ones for which analytic bounds could be obtained. In particular, it would be very interesting to apply cost-sensitivity to certifiable adversarial robustness \cite{cohen2019certified}, for which rigorous analytic results are possible. Lastly, it would also be interesting to extend beyond $\ell_p$ norm-based attacks, and consider more comprehensive attack sets \cite{gilmer2018motivating}.

\subsubsection*{Acknowledgments}
We would like to thank our colleagues at RAND with whom we had many fruitful discussions: Jair Aguirre, Caolionn O'Connnell, Edward Geist, Justin Grana, Christian Johnson, Osonde Osoba, Éder Sousa, Brian Vegetabile and Li Ang Zhang. This work was funded by RAND Project Air Force, contract number FA7014-16-D-1000. 

\bibliographystyle{JHEP}
\bibliography{refs}

\appendix

\section{\label{sec:uncalibrated} Additional results for uncalibrated attacks and predictions}
In Sec.~\ref{sec:experimental_results} we presented results for white-box attacks, with neither party calibrating. These attacks were both generated by and submitted to the ResNeXt-101 model of \cite{xie2018feature}. Results for black-box attacks may be obtained by submitting these same attacks to the other two  adversarially-robust models released by \cite{xie2018feature}, the ResNet-152 DeNoise and the ResNet-152 Baseline models. These are shown in Table~\ref{table:ImageNetresultsBlackBox}, again for $n_{\text{steps}} = 10$. These results are qualitatively similar to the white-box results, which demonstrates the transferability of adversarial attacks aimed at increasing the cost as well as the overall classification error.

\begin{table}[htbp]
   \caption{\label{table:ImageNetresultsBlackBox} Black-box adversarial attacks}
\centering
\begin{tabular}{ccccc}
    \toprule
    Attack Type & MP Acc. (\%) & MP Cost & MC Acc (\%) & MC Cost \\ \midrule
    \multicolumn{5}{c}{ResNet-152 Denoise} \\ \midrule
    \multicolumn{1}{l}{clean images} & $65.3 \pm 0.4$ & $7.22 \pm 0.32$ & $58.5 \pm 0.4$ & $2.68 \pm 0.14$   \\
    \multicolumn{1}{l}{random targets} & $55.4 \pm 0.4$ & $8.80 \pm 0.35$ & $45.0 \pm 0.4$ & $2.98 \pm 0.11$   \\
    \multicolumn{1}{l}{max cost targets} & $55.2 \pm 0.4$ & $8.87 \pm 0.35$ & $44.7 \pm 0.4$ & $3.07 \pm 0.12$   \\
    \multicolumn{1}{l}{maxi-min cost} & $58.7 \pm 0.4$ & $11.68 \pm 0.40$ & $49.5 \pm 0.4$ & $3.66 \pm 0.17$  \\ \midrule
    \multicolumn{5}{c}{ResNet-152 Baseline} \\ \midrule
    \multicolumn{1}{l}{clean images} & $62.3 \pm 0.4$ & $7.73 \pm 0.33$ & $55.7 \pm 0.4$ & $2.91 \pm 0.15$   \\
    \multicolumn{1}{l}{random targets} & $51.6 \pm 0.4$ & $9.47 \pm 0.36$ & $42.8 \pm 0.4$ & $3.24 \pm 0.13$   \\
    \multicolumn{1}{l}{max cost targets} & $51.7 \pm 0.4$ & $9.48 \pm 0.36$ & $42.8 \pm 0.4$ & $3.29 \pm 0.13$   \\
    \multicolumn{1}{l}{maxi-min cost} & $55.0 \pm 0.4$ & $12.03 \pm 0.41$ & $46.1 \pm 0.4$ & $3.99 \pm 0.18$ \\ \midrule
    \bottomrule
\end{tabular}
\end{table}

In addition to studying the transferability of attacks, we also investigated the dependence of the white-box attack efficacy on $n_{\text{step}}$. To this end, we generated 10,000 attacks with $n_{\text{step}}$ ranging from 10 to 1000. The results are plotted below in Fig.~\ref{fig:variable-nstep}, which shows the cost and accuracy for both types of predictions (maximum probability (MP) and minimum cost (MC)). The plots indicate that in many cases increasing number of steps to about 100 or 200 significantly improves the efficacy of the attack. In particular, larger values of $n_{\text{steps}}$ allows the targeted attacks with cost-sensitive targets to outperform the attacks with random targets in all cases. Additionally, with additional steps the efficacy of the maxi-min attack decreases relative to the other attacks \textit{against a minimum cost classifier}, as shown in the bottom-right figure. Against a maximum probability classifier, the maxi-min attack is far more effective at increasing the cost, as shown in the bottom-left figure. 

\begin{figure}[h!]
\centering
  \includegraphics[width=1.0\linewidth]{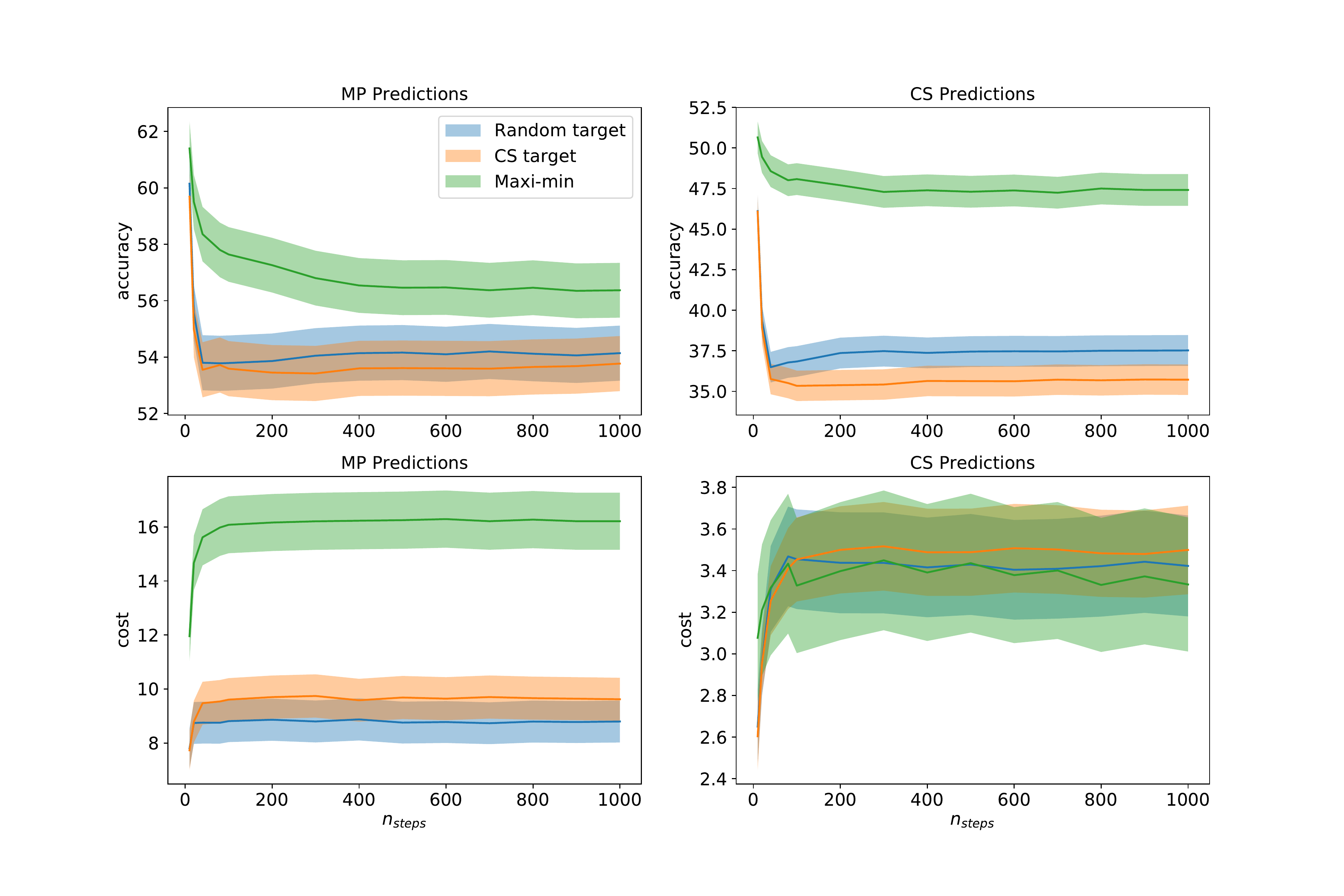}
  \captionof{figure}{The accuracy and cost as a function of $n_{\text{steps}}$ for the 3 different attacks considered here, and for both types of predictions: maximum probability (MP) and minimum cost (MC). Adversarial examples were generated for $n_{\text{steps}} = 1000$, with the output saved at intermediate values. Each curve represents an average over 10,000 adversarial examples, each for a different unperturbed ``clean'' image, and 95\% confidence intervals have been added around the mean. As is especially evident in the bottom-right plot, even with 10,000 images the confidence intervals are still quite large. Our analysis would benefit from larger samples sizes, which are unfortunately not practical given our computational resources and the time required to generate attacks with large values of $n_{\text{steps}}$.}
  \label{fig:variable-nstep}
\end{figure}  

\section{\label{sec:calibration_method} Temperature scaling calibration}
The calibration of neural networks was originally studied in \cite{niculescu2005predicting}. The issue was recently revisited for more modern architectures in \cite{guo2017calibration}, and we shall adopt their methodology.

The extent to which a classifier is well-calibrated may be measured by the Expected Calibration Error (ECE) \cite{naeini2015obtaining}, which is defined as
\begin{equation}
    \label{eq:ECE}
    \text{ECE} := \sum_{m=1}^M \frac{|B_m|}{n} \left| \text{acc}(B_m) - \text{conf}(B_m) \right| \,.
\end{equation}
Here, $B_m$ with $m=1,...,M$ represents a binning of predictions and $n$ is the total number of samples. Predictions are grouped into bin $B_m$ if their confidence (i.e. probability estimate $\max_A p(A|\bm{x})$) lies within the interval $\left(\frac{m-1}{M},\frac{m}{M}\right]$. Within each bin, the overall accuracy $\text{acc}(B_m)$ and average confidence $\text{conf}(B_m)$ are computed. An ECE of 0 indicates that the classifier is perfectly calibrated.

There are many techniques for calibrating a classifier. Perhaps the simplest is temperature scaling, in which the softmax operation relating the logits $z_A(\bm{x})$ to probabilities is modified via a temperature term as follows:
\begin{equation}
    p_T(A|\bm{x}) := \frac{\exp\left( z_A(\bm{x})/T \right)}{\sum_B \exp\left( z_B(\bm{x})/T \right)}
\end{equation}
For $T=1$, this reduces to the usual softmax operation. For $T > 1$, the probabilities are squeezed to become closer to one another, and for $T < 1$ the probabilities are pushed apart so that there is a wider disparity between them. The extreme limit of $T \rightarrow \infty$ corresponds to a uniform distribution, and the limit $T \rightarrow 0$ places all probability mass on the most probable label. An important property of temperature scaling is that it preserves the ordering of the probabilities. For example, the temperature scaling cannot change the sign of the relative log probabilities. Temperature scaling may be used to calibrate a classifier by using a separate validation set to find the optimal temperature $T_*$ which minimizes the ECE error, and then using this temperature to calibrate the probability estimates on the test set data. 

\begin{figure}[ht]
\centering
  \includegraphics[width=0.5\linewidth]{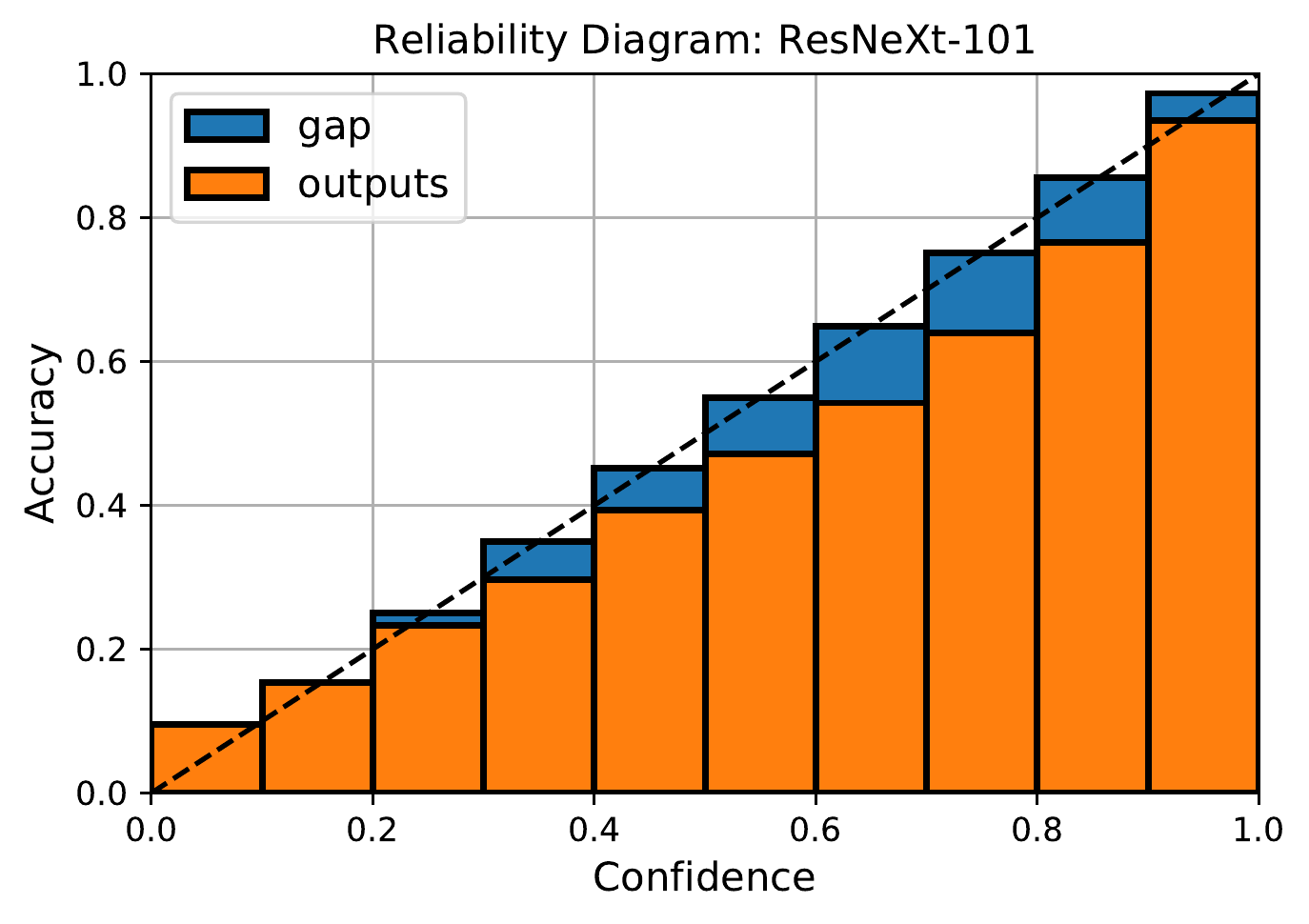}
  \captionof{figure}{Reliability diagram depicting the calibration of the ResNeXt-101 network when evaluated on the first 5000 images of the ImageNet validation set (representing 10\% of the full validation set). The gap represents the quantity within the absolute value sign in Eq.~\ref{eq:ECE}. The Expected Calibration Error (ECE) is 0.055, corresponding to a reasonably well-calibrated classifier.}
  \label{fig:reliability_diagram}
\end{figure}

Both the minimum cost prediction, Eq.~\ref{eq:MCprediction}, and the maximum minimum expected cost attack, Eq.~\ref{eq:maximin}, depend directly on the probability estimates $p(A|\bm{x})$, and so it is natural to wonder if calibration might significantly affect the results, for example by making the minimum cost predictions more robust, or the maximum minimum expected cost attack more effective. We investigated this issue for the white-box attacks in which both the attacking and defending network was the pre-trained ResNeXt-101 model of \cite{xie2018feature}. First, we evaluated the calibration of the ResNeXt-101 model, using 5000 images, representing 10\% of the full validation set. The ECE was found to be 0.055, representing a fairly well-calibrated classifier. To gain a better sense for the calibration, in Fig.~\ref{fig:reliability_diagram} below we plot the so-called reliability diagram \cite{guo2017calibration} showing $\text{conf}(B_m)$ vs. $\text{acc}(B_m)$.

The above reliability diagram and ECE value of 0.055 used the standard softmax operation, i.e. $T=1$. Allowing $T$ to vary, an optimal value of ECE $=0.016$ was found at the calibration temperature $T^* = 1.124$.

\section{\label{sec:calibration_scenarios} Calibration scenarios}
The calibration temperature of $T^* = 1.124$ found above could be used by the defender, the attacker, or both. The defender would be motivated to use calibrated probabilities so that their minimum cost predictions would be (hopefully) more accurate, and similarly the attacker would be motivated to use calibrated probabilities to generate more effective attacks. Thus, in the tables below we show results for the case where the defender calibrates but the attacker does not (Table~\ref{table:ImageNetresultsWhiteBox2}), the case where the defender does not calibrate but the attacker does (Table~\ref{table:ImageNetresultsWhiteBox3}), and the case in which both defender and attacker calibrate (Table~\ref{table:ImageNetresultsWhiteBox4}). The case in which neither party calibrates is covered above in Table~\ref{table:ImageNetresultsWhiteBox}. In all cases, the same calibration temperature $T^* = 1.124$ was used, and the results in the tables correspond to an average over the 45,000 validation images not used in the calibration step.

\begin{table}[h!]
   \caption{\label{table:ImageNetresultsWhiteBox2}Defender calibrates (white-box attack)}
\centering
\begin{tabular}{ccccc}
    \toprule
    Attack Type & MP Acc. (\%) & MP Cost & MC Acc (\%) & MC Cost \\ \midrule
    \multicolumn{5}{c}{ResNeXt-101} \\ \midrule
    \multicolumn{1}{l}{clean images} & $68.3 \pm 0.4$ & $6.50 \pm 0.32$ & $57.5 \pm 0.5$ & $2.35 \pm 0.11$   \\    
    \multicolumn{1}{l}{random targets} & $57.0 \pm 0.5$ & $8.44 \pm 0.36$ & $34.2 \pm 0.4$ & $3.21 \pm 0.05$   \\
    \multicolumn{1}{l}{max cost targets} & $56.9 \pm 0.5$ & $8.45 \pm 0.36$ & $33.7 \pm 0.4$ & $3.25 \pm 0.06$   \\
    \multicolumn{1}{l}{maxi-min cost} & $60.1 \pm 0.4$ & $13.94 \pm 0.46$ & $37.6 \pm 0.4$ & $3.38 \pm 0.10$  \\
    \midrule
\bottomrule
\end{tabular}
\end{table}

\begin{table}[h!]
   \caption{\label{table:ImageNetresultsWhiteBox3}Attacker calibrates (white-box attack)}
\centering
\begin{tabular}{ccccc}
    \toprule
    Attack Type & MP Acc. (\%) & MP Cost & MC Acc (\%) & MC Cost \\ \midrule
    \multicolumn{5}{c}{ResNeXt-101} \\ \midrule
    \multicolumn{1}{l}{clean images} & $68.3 \pm 0.4$ & $6.50 \pm 0.32$ & $57.5 \pm 0.5$ & $2.35 \pm 0.11$   \\    
    \multicolumn{1}{l}{random targets} & $57.0 \pm 0.5$ & $8.43 \pm 0.36$ & $41.3 \pm 0.5$ & $2.95 \pm 0.08$   \\
    \multicolumn{1}{l}{max cost targets} & $57.1 \pm 0.5$ & $8.45 \pm 0.36$ & $40.9 \pm 0.5$ & $3.00 \pm 0.09$   \\
    \multicolumn{1}{l}{maxi-min cost} & $61.3 \pm 0.5$ & $13.97 \pm 0.46$ & $55.0 \pm 0.5$ & $4.16 \pm 0.21$  \\
    \midrule
\bottomrule
\end{tabular}
\end{table}

\begin{table}[h!]
   \caption{\label{table:ImageNetresultsWhiteBox4}Both attacker and defender calibrate (white-box attack)}
\centering
\begin{tabular}{ccccc}
    \toprule
    Attack Type & MP Acc. (\%) & MP Cost & MC Acc (\%) & MC Cost \\ \midrule
    \multicolumn{5}{c}{ResNeXt-101} \\ \midrule
    \multicolumn{1}{l}{clean images} & $68.3 \pm 0.4$ & $6.50 \pm 0.32$ & $57.5 \pm 0.5$ & $2.35 \pm 0.11$   \\    
    \multicolumn{1}{l}{random targets} & $57.0 \pm 0.5$ & $8.43 \pm 0.36$ & $34.1 \pm 0.4$ & $3.22 \pm 0.06$   \\
    \multicolumn{1}{l}{max cost targets} & $57.1 \pm 0.5$ & $8.45 \pm 0.36$ & $33.7 \pm 0.4$ & $3.25 \pm 0.06$   \\
    \multicolumn{1}{l}{maxi-min cost} & $61.3 \pm 0.5$ & $13.97 \pm 0.46$ & $46.2 \pm 0.5$ & $3.39 \pm 0.14$  \\
    \midrule
\bottomrule
\end{tabular}
\end{table}

In discussing the results, let us first draw attention to the impact of calibration on the clean images. The maximum probability statistics are unaffected, which is to be expected since the temperature scaling method of calibration used here cannot change the maximum probability prediction.\footnote{The astute reader will have noticed that there are in fact slight differences between the MP results for clean un-calibrated and calibrated images. This are due to the fact that the averages computed in this section are over 45,000 images, as opposed to the 50,000 used in the previous section.} For the minimum cost predictions, the accuracy drops a non-trivial amount (from 61.2\% to 57.5\%) and the cost decreases slightly.

Moving next to consider the effect of calibration on the efficacy of the attacks, the results show that calibration (of either party) has a significant impact on the minimum cost predictions, but not on the maximum probability ones. In discussing the results, we will take the perspective of the defender, and assume that the attacker is held fixed. Consider first the case of an uncalibrated attacker. The results show that the two types of targeted attacks are much more effective against a calibrated minimum cost defender than an uncalibrated one. The accuracy decreases (from about 41\% to about 34\%) and the cost increases (from about 3 to about 3.2). Interestingly, the trend is reversed for the maxi-min attack. This attack is more effective against an uncalibrated minimum cost classifier (3.50 compared to 3.38 for a calibrated one). Thus, whether the defender should calibrate or not depends on the attack type.

Consider next the case in which the attacker calibrates. Once again, the maximum probability statistics are only very weakly affected by the defender's decision to calibrate. For the minimum cost predictions, it is again the case that the targeted attacks are more effective against a calibrated defender, whereas the maxi-min attack is rendered less effective by calibration. Here the distinction is even more pronounced than before. The cost for an uncalibrated minimum cost classifier is 4.16, and drops to 3.39 after calibration.

To summarize, calibration is important for minimum cost classifiers. A defender can reduce their vulnerability to a maxi-min attack designed to increase the expected cost by calibrating, and similarly an attacker can increase the effectiveness of the maxi-min attack against a minimum cost defender by calibrating. Against targeted attacks, however, calibration can decrease the defender's performance. In Sec.~\ref{sec:gametheory} we use game theory to conduct a more systematic analysis of the various strategies available to both the attacker and defender. 

\end{document}